\documentclass[10pt,twocolumn,letterpaper]{article}

\usepackage[pagenumbers]{cvpr} 

\usepackage{multirow}
\usepackage{colortbl}
\usepackage{arydshln}

\definecolor{firstcolor}{HTML}{BDE6CD}
\definecolor{secondcolor}{HTML}{E2EEBC}
\definecolor{thirdcolor}{HTML}{FFF8C5}

\newcommand{\fst}[1]{\cellcolor{firstcolor}\bfseries #1}
\newcommand{\snd}[1]{\cellcolor{secondcolor}#1}
\newcommand{\trd}[1]{\cellcolor{thirdcolor}#1}

\definecolor{cvprblue}{rgb}{0.21,0.49,0.74}
\usepackage[pagebackref,breaklinks,colorlinks,allcolors=cvprblue]{hyperref}

\def\paperID{****} 
\def\confName{CVPR}
\def\confYear{2026}

\title{PromptStereo: Zero-Shot Stereo Matching via Structure and Motion Prompts}

\author{Xianqi Wang$^{1}$, ~~Hao Yang$^{1}$, ~~Hangtian Wang$^{1}$\\~~Junda Cheng$^{1}$, ~~Gangwei Xu$^{1}$, ~~Min Lin$^{1}$, ~~Xin Yang$^{2,1}$\footnotemark[2]\\
[2mm]
$^1$Huazhong University of Science and Technology \quad $^2$Optics Valley Laboratory\\
{\tt\small \{xianqiw, haoyang2002, htwang, jundacheng, gwxu, minlin, xinyang2014\}@hust.edu.cn
}}

\begin{document}
\maketitle
\begin{abstract}
Modern stereo matching methods have leveraged monocular depth foundation models to achieve superior zero-shot generalization performance. However, most existing methods primarily focus on extracting robust features for cost volume construction or disparity initialization. At the same time, the iterative refinement stage, which is also crucial for zero-shot generalization, remains underexplored. Some methods treat monocular depth priors as guidance for iteration, but conventional GRU-based architectures struggle to exploit them due to the limited representation capacity. In this paper, we propose Prompt Recurrent Unit (PRU), a novel iterative refinement module based on the decoder of monocular depth foundation models. By integrating monocular structure and stereo motion cues as prompts into the decoder, PRU enriches the latent representations of monocular depth foundation models with absolute stereo-scale information while preserving their inherent monocular depth priors. Experiments demonstrate that our PromptStereo achieves state-of-the-art zero-shot generalization performance across multiple datasets, while maintaining comparable or faster inference speed. Our findings highlight prompt-guided iterative refinement as a promising direction for zero-shot stereo matching. Code: \textcolor{magenta}{https://github.com/Windsrain/PromptStereo}.
\end{abstract}

\footnotetext[2]{Corresponding author.}

\section{Introduction}
\label{sec:intro}

Stereo matching aims to estimate dense pixel-wise disparities from a pair of rectified images, providing essential depth information for 3D scene understanding. It plays a crucial role in applications such as autonomous driving.

\begin{figure}[t]
    \centering
    \includegraphics[width=1.0\linewidth]{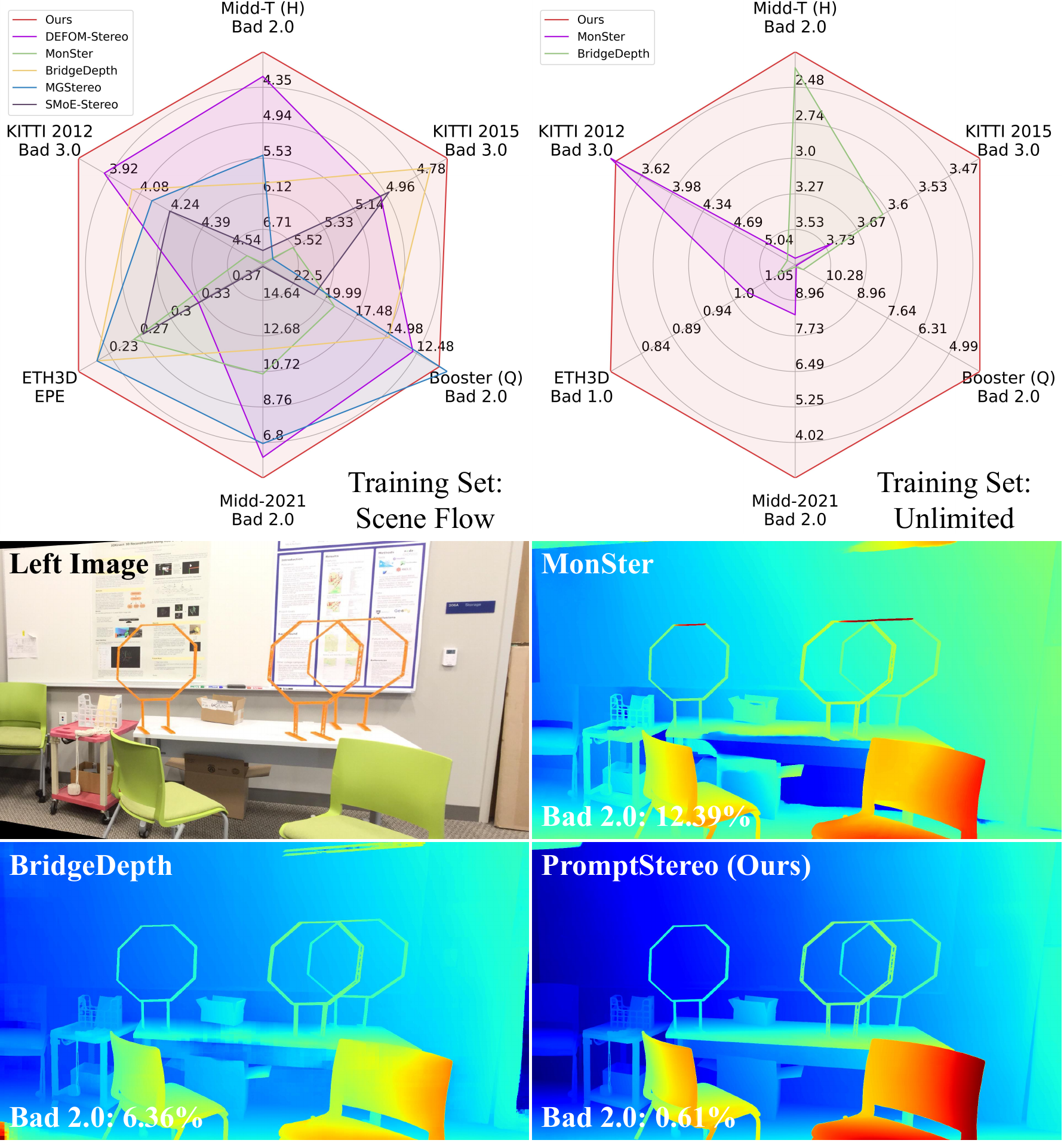}
    \caption{Comparison of zero-shot generalization. \textbf{Row 1:} Our PromptStereo outperforms previous methods. \textbf{Row 2-3:} A visualization example from Middlebury 2021~\cite{scharstein2014high}.}
    \label{fig:teaser}
    \vspace{-10px}
\end{figure}

With the rapid advancement of monocular depth foundation models~\cite{yang2024depth, bochkovskii2024depth, cheng2024adaptive, wang2025moge, he2024lotus, xu2025pixel}, zero-shot stereo matching has recently gained increasing attention. This growing interest comes from a clear shift in strategy: instead of designing complex modules to extract robust features from weak backbones~\cite{zhang2020domain, chuah2022itsa, zhang2022revisiting, chang2023domain, rao2023masked}, recent methods leverage strong monocular depth priors learned by powerful monocular models. By directly adapting pre-trained representations, these methods~\cite{jiang2025defom, wen2025foundationstereo, cheng2025monster, bartolomei2025stereo, guan2025bridgedepth, yao2025diving, wang2025learning} achieve simpler designs and stronger generalization performance.

However, the iterative refinement stage has been relatively underexplored. Recent methods utilize monocular depth foundation models to extract depth-aware features, construct robust cost volumes, and obtain accurate initial disparities for iteration. Some methods further introduce monocular depth priors to guide GRU~\cite{cho2014learning}, a recurrent unit popularized by RAFT-Stereo~\cite{lipson2021raft}. Such recurrent units are increasingly limited by their restricted capacity and scalability. First, GRU is independent of vision foundation models and must be trained from scratch, preventing it from inheriting strong priors. As a result, it's hard to gain visual understanding capacity. Second, GRU constrains its hidden states within a narrow range, making it challenging to handle extreme disparity variations or complex geometric structures. Finally, GRU fuses inputs and hidden states through direct convolutions, which not only distort the original state information but also compress external inputs, leading to ambiguous guidance.

To overcome these limitations, we propose Prompt Recurrent Unit (PRU), a novel recurrent unit that rethinks iterative refinement from the perspective of vision foundation models. While past methods rely on multi-level GRU for coarse-to-fine disparity update, we find that the decoder of monocular depth foundation models, such as DPT~\cite{ranftl2021vision}, also follows a similar multi-resolution refinement architecture. Therefore, PRU adopts this decoder architecture as a unified multi-level recurrent unit, naturally inheriting rich monocular depth priors. To integrate additional inputs without disrupting these priors, we propose Structure Prompt (SP) and Motion Prompt (MP), which prompt monocular structure and stereo motion cues into PRU. Compared to previous prompt-based attempts in monocular depth estimation~\cite{lin2025prompting, wang2025depth}, stereo matching enables richer prompts without the need for external sensors. To enable iterative refinement similar to GRU, we design a simple yet effective update strategy for PRU, which removes range constraints on hidden states, bringing more straightforward and flexible iterative refinement. Furthermore, Affine-Invariant Fusion (AIF) merges the initial disparity and monocular depth under a normalized scale, resulting in faster convergence and better geometric consistency. 

By replacing GRU with PRU in existing methods, we propose PromptStereo. As shown in Fig. ~\ref {fig:teaser}, PromptStereo achieves state-of-the-art zero-shot generalization performance across multiple datasets. The performance on unlimited training sets demonstrates the superior representation capacity and scalability of PRU. Subsequent experiments also prove that PromptStereo maintains comparable or faster inference speed.

Our contributions can be summarized as follows:

\begin{itemize}
\item We propose Prompt Recurrent Unit (PRU), a novel recurrent unit built upon the decoder of monocular depth foundation models. It directly inherits monocular depth priors and holds better representation capacity and scalability than GRU.

\item We propose Structure Prompt (SP) and Motion Prompt (MP), which prompt monocular structure and stereo motion cues into PRU, avoiding distorted state information and ambiguous guidance.

\item We propose a simple yet effective update strategy for PRU and Affine-Invariant Fusion (AIF) for the initial disparity, both beneficial to iterative refinement.

\item Our PromptStereo achieves state-of-the-art zero-shot generalization performance across various datasets, while maintaining comparable or faster inference speed.

\end{itemize}
\section{Related Work}
\label{sec:work}

\textbf{Deep Stereo Matching.} Deep learning has advanced stereo matching by replacing hand-crafted features with learnable representations. Since DispNet~\cite{mayer2016large} and GC-Net~\cite{kendall2017end}, end-to-end architectures have been widely adopted in this field. These aggregation-based methods~\cite{chang2018pyramid, guo2019group, xu2020aanet, shen2021cfnet, xu2022attention, cheng2022region, cheng2024coatrsnet} follow a four-step pipeline of feature extraction, cost volume construction, cost aggregation, and disparity regression, progressively improving accuracy and efficiency. After that, iterative-based methods have become the mainstream of research. These methods~\cite{lipson2021raft, li2022practical, xu2023iterative, zhao2023high, wang2024selective, chen2024mocha, chen2024motif, chen2025feature, li2025global} utilize local cost volume lookup and iterative refinement to replace cost aggregation, achieving more accurate and efficient high-resolution inference. More recently, many methods~\cite{jiang2025defom, wen2025foundationstereo, cheng2025monster, bartolomei2025stereo, guan2025bridgedepth, yao2025diving, wang2025learning, cheng2025monster++} have leveraged vision foundation models, particularly monocular depth foundation models, to enhance model performance. They utilize these models to extract robust features~\cite{wen2025foundationstereo, wang2025learning}, construct cost volumes~\cite{bartolomei2025stereo}, and treat monocular depth priors as guidance~\cite{jiang2025defom, cheng2025monster, guan2025bridgedepth, yao2025diving}, achieving superior performance.

\textbf{Zero-Shot Stereo Matching.} The demand for zero-shot stereo matching is gradually increasing, aiming to enable models to adapt to various scenarios. Initially, most methods introduce new modules for zero-shot generalization. DSMNet~\cite{zhang2020domain} proposes domain normalization layers and non-local graph-based filters to learn domain-invariant features. ITSA~\cite{chuah2022itsa} borrows the concept of shortcut learning to avoid domain-specific shortcuts during training. MRL-Stereo~\cite{rao2023masked} utilizes mask representation learning to learn domain-invariant structure information. With the development of visual foundation models, many methods~\cite{liu2022graftnet, wang2025zerostereo, wen2025foundationstereo, bartolomei2025stereo, wang2025learning} directly utilize these models to improve zero-shot generalization performance. GraftNet~\cite{liu2022graftnet} firstly incorporates pre-trained features to obtain task-related information. FoundationStereo~\cite{wen2025foundationstereo} introduces a large-scale dataset to help zero-shot generalization. Stereo Anywhere~\cite{bartolomei2025stereo} constructs a normal cost volume and proposes novel volume augmentations. SMoE-Stereo~\cite{wang2025learning} dynamically selects suitable features from several vision foundation models. MGStereo~\cite{yao2025diving} explores the fusion of monoculor priors via ordering local fusion and registered global fusion. Some methods~\cite{jiang2025defom, cheng2025monster} also achieve strong generalization results, even though their original purpose is in-domain performance. However, these methods overlook the limited representation capacity and scalability of GRU, which constrains their generalization ability. Our PromptStereo is the first to enable the iterative refinement to inherit monocular depth priors, achieving impressive zero-shot generalization performance across various datasets.
\begin{figure*}[t]
    \centering
    \includegraphics[width=1.0\textwidth]{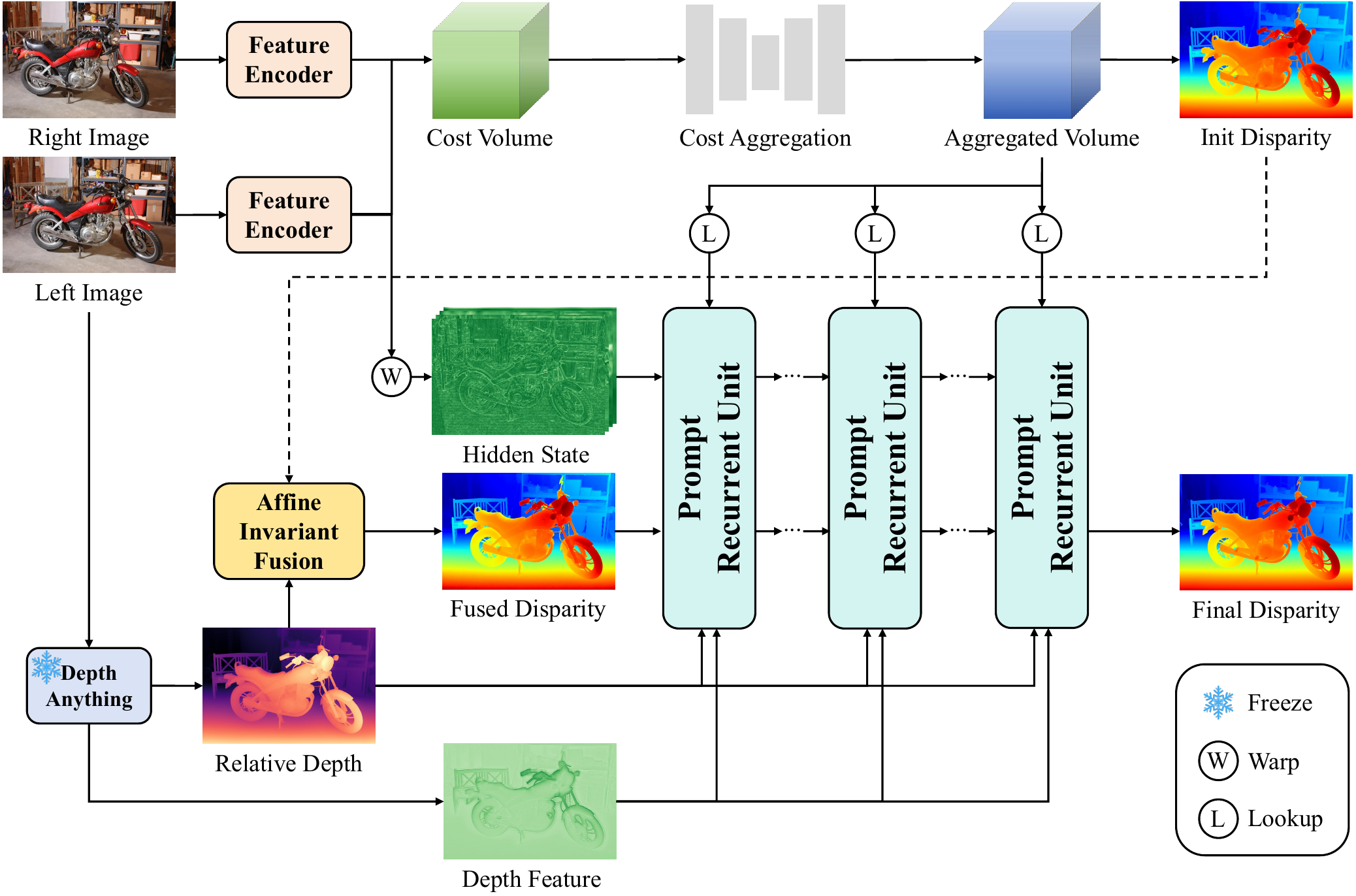}
    \caption{Overview of PromptStereo. It builds upon MonSter~\cite{cheng2025monster}, sharing the same feature extraction and cost volume construction. The initial disparity and relative depth are first fused via Affine-Invariant Fusion (Sec. ~\ref{subsec:fusion}). The fused disparity is then refined iteratively by our Prompt Recurrent Unit (Sec. ~\ref{subsec:unit}), which replaces the GRU used in previous methods. It enables effective, prompt-guided iterative refinement, yielding the final high-quality disparity.}
    \label{fig:overview}
    \vspace{-10px}
\end{figure*}

\section{Method}
\label{sec:method}

In this section, we present an overview of our PromptStereo (Fig. ~\ref{fig:overview}) and details of our proposed modules, including Affine-Invariant Fusion and Prompt Recurrent Unit.

\subsection{Feature Extraction}

We take MonSter~\cite{cheng2025monster} as our baseline. Given a stereo image pair $\mathbf{I}_L, \mathbf{I}_R \in \mathbb{R}^{3 \times H \times W}$, the monocular feature branch utilizes a pre-trained Depth Anything V2~\cite{yang2024depth} to extract a relative depth $\mathbf{d}_M \in \mathbb{R}^{1 \times H \times W}$ and the depth feature before the last layer $\mathbf{F}_M \in \mathbb{R}^{C \times \frac{H}{4} \times \frac{W}{4}}$. This branch is frozen during training to maintain robust monocular depth priors. The stereo feature branch utilizes MonSter's feature encoder, which consists of a pre-trained Depth Anything V2 and a feature transfer network, to extract multi-level stereo features $\mathbf{F}_L^i, \mathbf{F}_R^i \in \mathbb{R}^{C_i \times \frac{H}{2^{i + 2}} \times \frac{W}{2^{i + 2}}} \, (i = 0, 1, 2, 3)$. We follow the original setting to freeze the DINOv2~\cite{oquab2023dinov2} encoder and keep other networks trainable. We do not use MonSter's checkpoint initialization during training to ensure a fair comparison.

\subsection{Cost Volume Construction}

We follow the cost volume construction strategy of MonSter~\cite{cheng2025monster}, derived from IGEV-Stereo~\cite{xu2023iterative}. Specifically, we first use $\mathbf{F}_L^0, \mathbf{F}_R^0$ to build a group-wise correlation volume~\cite{guo2019group} $\mathbf{V}_G \in \mathbb{R}^{G \times \frac{D}{4} \times \frac{H}{4} \times \frac{W}{4}}$, and then use a lightweight 3D aggregation network~\cite{bangunharcana2021correlate} to filter it, yielding the geometry encoding volume $\mathbf{V}_E \in \mathbb{R}^{G \times \frac{D}{4} \times \frac{H}{4} \times \frac{W}{4}}$. In parallel, an all-pair correlation volume~\cite{lipson2021raft} $\mathbf{V}_A \in \mathbb{R}^{1 \times \frac{W}{4} \times \frac{H}{4} \times \frac{W}{4}}$ is computed to capture global feature similarity. Finally, volumes are pooled to form multi-level volume pyramids, which are collectively referred to as the combined geometry encoding volume $\mathbf{V}_C$. The cost volume construction is formulated as:
\begin{equation}
    \begin{aligned}
        \mathbf{V}_G (g, d, h, w) & = \frac{G}{C_0} \cdot \langle \mathbf{F}_{L,g}^0 (h, w), \mathbf{F}_{R,g}^0 (h, w - d) \rangle \\
        \mathbf{V}_A (w', h, w) & = \langle \mathbf{F}_L^0 (h, w), \mathbf{F}_R^0 (h, w') \rangle
    \end{aligned}
\end{equation}
where $\langle \cdot \rangle$ is the inner product operation; g is the group index; d is the disparity index; $\mathbf{F}_{L,g}^0$ is the g-th group of feature along the channel dimension, and $\mathbf{F}_{R,g}^0$ is the same.

\begin{figure*}[t]
    \centering
    \includegraphics[width=1.0\textwidth]{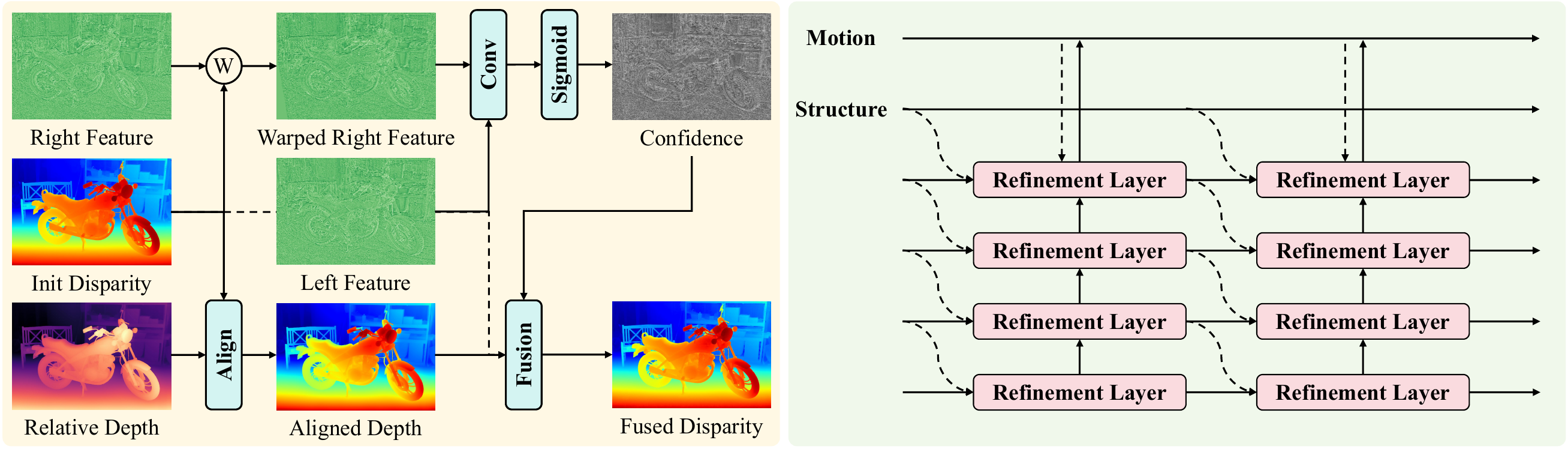}
    \caption{Overview of proposed modules. Left: Affine-Invariant Fusion. Right: Prompt Recurrent Unit.}
    \label{fig:module}
    \vspace{-10px}
\end{figure*}

\subsection{Affine-Invariant Fusion}
\label{subsec:fusion}

While the cost volume construction can provide a reasonable initial disparity, it lacks global geometric consistency due to its limited disparity range. On the other hand, the relative depth offers a strong structure prior, but suffers from affine ambiguity in scale and shift. To obtain a more reliable initialization, we propose Affine-Invariant Fusion (Fig. ~\ref{fig:module}).

Given the relative depth $\mathbf{d}_M$ and the initial disparity $\mathbf{d}_0$ regressed from $\mathbf{V}_E$, we normalize them in an affine-invariant manner to obtain $\hat{\mathbf{d}}_M, \hat{\mathbf{d}}_0$: 
\begin{equation}
    \begin{aligned}
        t(\mathbf{d}) & = \text{median}(\mathbf{d}) \\
        s(\mathbf{d}) & = \frac{1}{N} \sum_{i = 1}^N | \mathbf{d} - t(\mathbf{d}) | \\
        \hat{\mathbf{d}} & = \frac{\mathbf{d} - t(\mathbf{d})}{s(\mathbf{d})}
    \end{aligned}
\end{equation}
where $t(\mathbf{d}), s(\mathbf{d})$ are used to normalize $\mathbf{d}$. This operation is commonly used in monocular depth estimation as an affine-invariant loss supervision ~\cite{ranftl2020towards, yang2024depth}. We adopt it for the initial alignment because the operation is simple and robust. Although the alignment may not be perfectly accurate, the subsequent iterative refinement can effectively correct any residual misalignment. Then, we project $\hat{\mathbf{d}}_M$ into the disparity space:
\begin{equation}
    \mathbf{d}_M' = s(\mathbf{d}_0) \cdot \hat{\mathbf{d}}_M + t(\mathbf{d}_0)
\end{equation}

Next, $\mathbf{d}_0$ is used to warp the right feature $\mathbf{F}_R^0$, which is then concatenated with the left feature $\mathbf{F}_L^0$. A convolutional block followed by a sigmoid function produces the confidence map $\mathbf{c}$, representing the reliability of $\mathbf{d}_0$. The fused disparity $\mathbf{d}_F$ is obtained as follows:
\begin{equation}
    \mathbf{d}_F = \mathbf{c} \odot \mathbf{d}_0 + (1 - \mathbf{c}) \odot \mathbf{d}_M'
\end{equation}
where $\odot$ is the element-wise product operation; $\mathbf{d}_F$ serves as the starting point for the subsequent iterative refinement.

\subsection{Prompt Recurrent Unit}
\label{subsec:unit}

To overcome the limitations of GRU, we propose Prompt Recurrent Unit (PRU), a novel recurrent unit built upon the decoder of monocular depth foundation models.

\textbf{State Initialization.} We adopt the refinement layers of the DPT~\cite{ranftl2021vision} decoder from Depth Anything V2~\cite{yang2024depth}, which naturally form a multi-resolution architecture. These layers are initialized with pre-trained weights, allowing PRU to inherit monocular depth priors while maintaining scalability for stereo refinement. For the hidden state initialization, we concatenate the left features $\mathbf{F}_L^i$ and the warped right features $\mathbf{F}_R^i$ using $\mathbf{d}_0$, and feed them into convolutional blocks to generate the initial hidden state $\mathbf{h}_0^i \, (i = 0, 1, 2, 3)$. Compared to traditional GRU initialization, which only uses left features, our initialization encourages hidden states to learn stereo correspondences at an earlier stage.

\textbf{Structure Prompt.} The frozen monocular depth feature $\mathbf{F}_M$ and relative depth $\mathbf{d}_M$ provide robust global geometric information, but would distort the stereo disparity range if using them directly. To address this, we normalize the current disparity $\mathbf{d}_k$ to $\hat{\mathbf{d}}_k$ as Sec. ~\ref{subsec:fusion} does, and compute difference with the normalized relative depth $\hat{\mathbf{d}}_M$:
\begin{equation}
    \mathbf{D} = | \hat{\mathbf{d}}_k - \hat{\mathbf{d}}_M |
\end{equation}
where $\mathbf{D}$ is the difference. This difference captures affine-invariant geometric discrepancies between monocular and stereo predictions, highlighting regions where geometric alignment is inconsistent. Such information serves as an effective structure-aware cue to guide iterative refinement. Structure Prompt (SP) is formulated as follows:
\begin{equation}
    \begin{aligned}
        \mathbf{P}_S & = \text{Encoder}(\mathbf{F}_M, \mathbf{D}) \\
        \mathbf{h} & = \mathbf{h} + \text{ConvBlock}(\mathbf{P}_S)
    \end{aligned}
\end{equation}
where $Encoder$ is the structure encoder with the same architecture as previous methods~\cite{lipson2021raft, xu2023iterative}; $ConvBlock$ is a convolutional block; $\mathbf{h}$ is the intermediate hidden state. This residual addition serves as a feature-level prompt, guiding the hidden state without disrupting the inherited priors.

\textbf{Motion Prompt.} Motion information encodes stereo-related cues such as correlation and disparity. We adopt a motion encoder similar to previous methods~\cite{lipson2021raft, xu2023iterative}, and define our Motion Prompt (MP) as follows:
\begin{equation}
    \begin{aligned}
        \mathbf{P}_M^k & = \text{Encoder}(\mathbf{V}_k, \mathbf{d}_k) \\
        \mathbf{h} & = \mathbf{h} + \text{ConvBlock}(\mathbf{P}_M^k)
    \end{aligned}
\end{equation}
where $Encoder$ is the motion encoder; $\mathbf{V}_k$ is the local cost volume. This residual prompting allows the model to integrate stereo motion cues adaptively.

\textbf{Update Strategy.} Unlike GRU, which updates the hidden state through both reset and update gates, we design a simple and efficient update strategy for PRU. The following equations summarize PRU's internal operations, including prompt encoding, prompt fusion, and state update:
\begin{equation}
    \begin{aligned}
        \mathbf{z}_k & = \begin{cases}
            \sigma (\text{ConvBlock}([\mathbf{h}^i_k, \mathbf{h}^{i - 1}_k])), & i > 0 \\
            \sigma (\text{ConvBlock}([\mathbf{h}^i_k, \mathbf{P}_S, \mathbf{P}_M^k])), & i = 0
        \end{cases} \\
        \hat{\mathbf{h}}_k^i & = {\mathbf{h}}_k^i + \text{ResBlock}(\mathbf{h}^{i + 1}_{k + 1}), \mspace{87mu} i < 3 \\
        \hat{\mathbf{h}}_k^i & = \text{ResBlock}(\hat{\mathbf{h}}_k^i) \\
        \hat{\mathbf{h}}_k^i & = \hat{\mathbf{h}}_k^i + \text{ConvBlock}(\mathbf{P}_S), \hspace{5em} i = 0 \\
        \hat{\mathbf{h}}_k^i & = \hat{\mathbf{h}}_k^i + \text{ConvBlock}(\mathbf{P}_M^k), \mspace{85mu} i = 0 \\
        \hat{\mathbf{h}}_k^i & = \text{Conv}_{1 \times 1}(\hat{\mathbf{h}}_k^i) \\
        \mathbf{h}_{k + 1}^i & = (1 - \mathbf{z}_k) \odot \mathbf{h}_k^i + \mathbf{z}_k \odot \hat{\mathbf{h}}_k^i \\
        \mathbf{d}_{k + 1} & = \mathbf{d}_{k} + \text{ConvBlock}(\mathbf{h}_{k + 1}^i)
    \end{aligned}
    \label{equ:update}
\end{equation}
where $[\cdot]$ is the concatenation operation; $\sigma$ is the sigmoid function; $ResBlock$ is a residual convolutional block. Due to the multi-resolution architecture, some operations are applied conditionally. For instance, prompts are only injected at the highest resolution. As shown in Equ. ~\ref{equ:update}, PRU does not constrain the hidden state as GRU does, providing a more flexible representation space. Moreover, we remove the reset gate and only use the update gate $\mathbf{z}_k$. $\mathbf{z}_k$ is computed using the higher-resolution hidden state, while the input to the current hidden state comes solely from the lower-resolution hidden state. Unlike GRU, which updates the medium-resolution hidden state using both, this design reduces computational complexity and improves both training efficiency and inference speed.

\subsection{Loss Function}

We follow IGEV-Stereo~\cite{xu2023iterative} to build the loss function:
\begin{equation}
    \mathcal{L} = ||\mathbf{d}_0 - \mathbf{d}_{gt}||_{smooth} + \sum_{k=1}^K \gamma ^{K - i} ||\mathbf{d}_k - \mathbf{d}_{gt}||_1
\end{equation}
where $|| \cdot ||_{smooth}$ is the smooth L1 loss; $|| \cdot ||_1$ is the L1 loss; $\mathbf{d}_0$ is the initial disparity; $\mathbf{d}_{gt}$ is the ground-truth disparity; $\mathbf{d}_k$ is the predicted disparity each iteration; $K$ is the iteration number; $\gamma = 0.9$.
\section{Experiment}
\label{sec:experiment}

In this section, we present our implementation details, evaluation datasets, evaluation metrics, and experiment results. We also refer readers to our supplementary materials for extra experiments and comparisons with specific methods not included in the main text. These methods include the use of special data augmentations, training strategies, and other techniques, which make a fair comparison infeasible.

\subsection{Implementation Detail}

We implement our PromptStereo using PyTorch and train it on 4 NVIDIA RTX 4090 GPUs. We use MonSter~\cite{cheng2025monster} as our baseline. For all training, we use the AdamW optimizer and the one-cycle learning rate scheduler with a learning rate of 2e-4. For Scene Flow evaluation, we train it on Scene Flow~\cite{mayer2016large} with a batch size of 8 for 200k steps from scratch. For unlimited evaluation, we train it on a mixed dataset consisting of FoundationStereo~\cite{wen2025foundationstereo}, CREStereo~\cite{li2022practical}, Falling Things~\cite{tremblay2018falling}, Scene Flow~\cite{mayer2016large}, and Virtual KITTI 2~\cite{cabon2020virtual} from the Scene Flow checkpoint. We apply standard data augmentations following RAFT-Stereo~\cite{lipson2021raft} with a crop size of $384 \times 768$. We use 16 iterations during training and 32 iterations during inference.

\subsection{Evaluation Dataset and Metric}

\textbf{Evaluation Datasets.} We use 5 datasets for basic evaluation. \textbf{KITTI 2012}~\cite{geiger2012we} and \textbf{KITTI 2015}~\cite{menze2015object} are datasets for outdoor driving scenarios, containing 194 and 200 training pairs with sparse LiDAR ground truth, respectively. \textbf{Middlebury V3 Benchmark Training Set (Midd-T)}~\cite{scharstein2014high} contains 15 training pairs with high-quality disparity ground truth for high-resolution indoor scenarios. \textbf{Middlebury 2021 (Midd-2021)}~\cite{scharstein2014high} also includes 24 training pairs for such scenarios obtained with a mobile device on a robot arm. \textbf{ETH3D}~\cite{schops2017multi} contains 27 training grayscale stereo image pairs with semi-dense disparity ground truth for low-resolution indoor/outdoor scenarios. Besides, we consider 2 datasets for advanced evaluation, covering challenging scenarios. \textbf{DrivingStereo}~\cite{yang2019drivingstereo} comprises 4 subsets with varying challenging weather conditions for outdoor driving scenarios. Each subset has 500 training pairs with sparse LiDAR ground truth. Booster~\cite{ramirez2022open} contains 228 training pairs with high-quality disparity ground truth for high-resolution indoor scenarios. It provides specular and transparent surfaces, which are the leading cause of failures in modern stereo matching.

\textbf{Evaluation Metrics.} We evaluate our PromptStereo using two standard metrics: EPE, which computes the per-pixel absolute disparity error, and Bad $\tau$, which calculates the percentage of pixels with a disparity error greater than $\tau$ pixels. If possible, we compute all pixels, referred to as \textit{All}, and non-occluded pixels, referred to as \textit{Noc}.

\begin{table*}
    \centering
    \resizebox{\textwidth}{!}{
    \begin{tabular}{l|cccccccccccccccccccc}
    \hline
    \multirow{3}{*}{Method} & \multicolumn{4}{c|}{KITTI 2012} & \multicolumn{4}{c|}{KITTI 2015} & \multicolumn{4}{c|}{Midd-T (H)} & \multicolumn{4}{c|}{Midd-2021} & \multicolumn{4}{c}{ETH3D} \\ \cline{2-21} 
     & \multicolumn{2}{c}{EPE} & \multicolumn{2}{c|}{Bad 3.0} & \multicolumn{2}{c}{EPE} & \multicolumn{2}{c|}{Bad 3.0} & \multicolumn{2}{c}{EPE} & \multicolumn{2}{c|}{Bad 2.0} & \multicolumn{2}{c}{EPE} & \multicolumn{2}{c|}{Bad 2.0} & \multicolumn{2}{c}{EPE} & \multicolumn{2}{c}{Bad 1.0} \\
     & All & Noc & All & \multicolumn{1}{c|}{Noc} & All & Noc & All & \multicolumn{1}{c|}{Noc} & All & Noc & All & \multicolumn{1}{c|}{Noc} & All & Noc & All & \multicolumn{1}{c|}{Noc} & All & Noc & All & Noc \\ \hline
    \multicolumn{1}{l}{Training Set} & \multicolumn{20}{c}{Scene Flow} \\ \hline
    RAFT-Stereo~\cite{lipson2021raft} & 0.90 & 0.83 & 4.34 & \multicolumn{1}{c|}{3.86} & 1.13 & 1.11 & 5.68 & \multicolumn{1}{c|}{5.45} & 1.40 & 1.03 & 11.07 & \multicolumn{1}{c|}{8.41} & 1.40 & 0.93 & 11.11 & \multicolumn{1}{c|}{6.38} & \trd 0.27 & \trd 0.24 & 2.61 & 2.29 \\
    IGEV-Stereo~\cite{xu2023iterative} & 1.03 & 0.93 & 5.13 & \multicolumn{1}{c|}{4.50} & 1.21 & 1.17 & 6.03 & \multicolumn{1}{c|}{5.79} & 1.48 & 0.88 & 9.95 & \multicolumn{1}{c|}{7.09} & 1.33 & 0.86 & 10.00 & \multicolumn{1}{c|}{\trd 5.35} & 0.33 & 0.29 & 4.05 & 3.61 \\
    DEFOM-Stereo~\cite{jiang2025defom} & \snd 0.83 & \trd 0.78 & \snd 3.90 & \multicolumn{1}{c|}{\snd 3.52} & \fst 1.06 & \fst 1.04 & 4.99 & \multicolumn{1}{c|}{\trd 4.76} & \snd 0.94 & \snd 0.65 & \snd 6.77 & \multicolumn{1}{c|}{\snd 4.17} & \snd 1.24 & \snd 0.76 & \snd 8.62 & \multicolumn{1}{c|}{\snd 4.98} & 0.35 & 0.33 & 2.40 & 2.16 \\
    MonSter\footnotemark[1]~\cite{cheng2025monster} & 0.93 & 0.88 & 4.62 & \multicolumn{1}{c|}{4.12} & 1.17 & 1.15 & 5.52 & \multicolumn{1}{c|}{5.32} & 1.04 & 0.94 & 8.97 & \multicolumn{1}{c|}{7.27} & 1.70 & 1.21 & 15.55 & \multicolumn{1}{c|}{10.58} & 0.35 & 0.26 & 3.20 & 2.86 \\
    BridgeDepth~\cite{guan2025bridgedepth} & \snd 0.83 & \snd 0.77 & \trd 4.04 & \multicolumn{1}{c|}{\trd 3.63} & \snd 1.09 & \trd 1.07 & \snd 4.69 & \multicolumn{1}{c|}{\snd 4.52} & \fst 0.92 & \trd 0.74 & 7.84 & \multicolumn{1}{c|}{5.94} & 1.91 & 1.56 & 15.92 & \multicolumn{1}{c|}{11.93} & \fst 0.23 & \snd 0.22 & \fst 1.26 & \fst 1.14 \\
    GREAT-IGEV~\cite{li2025global} & 1.02 & 0.93 & 5.31 & \multicolumn{1}{c|}{4.71} & 1.16 & 1.14 & 5.88 & \multicolumn{1}{c|}{5.64} & 1.41 & 0.86 & 9.23 & \multicolumn{1}{c|}{6.17} & \trd 1.25 & 0.95 & \trd 9.46 & \multicolumn{1}{c|}{5.37} & 0.37 & 0.28 & 4.47 & 3.80 \\
    MGStereo~\cite{yao2025diving} & \trd 0.86 & 0.80 & 4.14 & \multicolumn{1}{c|}{3.71} & \trd 1.12 & 1.10 & 5.64 & \multicolumn{1}{c|}{5.40} & 1.12 & 0.81 & 8.37 & \multicolumn{1}{c|}{\trd 5.48} & 1.52 & \trd 0.92 & 11.55 & \multicolumn{1}{c|}{6.74} & \snd 0.25 & \snd 0.22 & \trd 1.88 & \trd 1.59 \\
    SMoE-Stereo~\cite{wang2025learning} & \snd 0.83 & \snd 0.77 & 4.23 & \multicolumn{1}{c|}{3.76} & \fst 1.06 & \fst 1.04 & \trd 4.94 & \multicolumn{1}{c|}{4.77} & 1.54 & 1.29 & \trd 7.38 & \multicolumn{1}{c|}{7.06} & 4.61 & 3.50 & 21.02 & \multicolumn{1}{c|}{16.51} & \trd 0.27 & \snd 0.22 & 2.44 & 2.08 \\
    PromptStereo (Ours) & \fst 0.79 & \fst 0.73 & \fst 3.77 & \multicolumn{1}{c|}{\fst 3.32} & \snd 1.09 & \snd 1.06 & \fst 4.59 & \multicolumn{1}{c|}{\fst 4.39} & \trd 0.96 & \fst 0.60 & \fst 6.03 & \multicolumn{1}{c|}{\fst 3.76} & \fst 0.95 & \fst 0.75 & \fst 8.26 & \multicolumn{1}{c|}{\fst 4.84} & \fst 0.23 & \fst 0.20 & \snd 1.56 & \snd 1.30 \\ \hline
    \multicolumn{1}{l}{Training Set} & \multicolumn{20}{c}{Unlimited} \\ \hline
    FoundationStereo\footnotemark[2]~\cite{wen2025foundationstereo} & 0.68 & 0.63 & 3.00 & \multicolumn{1}{c|}{2.70} & 0.87 & 0.85 & 3.10 & \multicolumn{1}{c|}{2.98} & 0.66 & 0.38 & 3.11 & \multicolumn{1}{c|}{1.18} & 1.00 & 0.73 & 7.14 & \multicolumn{1}{c|}{3.65} & 0.17 & 0.15 & 0.67 & 0.49 \\ \hdashline
    MonSter~\cite{cheng2025monster} & \snd 0.73 & \snd 0.68 & \fst 3.27 & \multicolumn{1}{c|}{\fst 2.96} & \snd 0.91 & \trd 0.90 & \trd 3.72 & \multicolumn{1}{c|}{\trd 3.55} & \trd 0.78 & \trd 0.59 & \trd 5.51 & \multicolumn{1}{c|}{\trd 3.75} & \trd 1.87 & \trd 1.50 & \snd 12.43 & \multicolumn{1}{c|}{\snd 8.46} & 
    \snd 0.19 & \snd 0.16 & \trd 1.25 & \snd 1.02 \\
    BridgeDepth~\cite{guan2025bridgedepth} & \fst 0.70 & \fst 0.65 & \trd 5.31 & \multicolumn{1}{c|}{\trd 4.71} & \snd 0.91 & \snd 0.89 & \snd 3.61 & \multicolumn{1}{c|}{\snd 3.48} & \snd 0.74 & \snd 0.57 & \fst 3.36 & \multicolumn{1}{c|}{\snd 2.33} & \snd 1.52 & \snd 1.23 & \trd 13.66 & \multicolumn{1}{c|}{\trd 10.13} & \snd 0.19 & \trd 0.18 & \snd 1.22 & \trd 1.07 \\
    PromptStereo (Ours) & \fst 0.70 & \fst 0.65 & \snd 3.33 & \multicolumn{1}{c|}{\snd 3.05} & \fst 0.88 & \fst 0.87 & \fst 3.40 & \multicolumn{1}{c|}{\fst 3.27} & \fst 0.59 & \fst 0.44 & \snd 3.90 & \multicolumn{1}{c|}{\fst 2.21} & \fst 0.78 & \fst 0.61 & \fst 5.97 & \multicolumn{1}{c|}{\fst 2.78} & \fst 0.16 & \fst 0.15 & \fst 0.97 & \fst 0.79 \\ \hline
    \end{tabular}
    }
    \vspace{-5px}
    \caption{Zero-shot generalization basic benchmark. MonSter$^*$ is re-trained with the official code. FoundationStereo$^{\dagger}$ is reported for reference only, as it requires significantly large batch sizes and model parameters to train, making a fair comparison infeasible.}
    \label{tab:basic}
    \vspace{-5px}
\end{table*}

\begin{figure*}[t]
    \centering
    \includegraphics[width=1.0\textwidth]{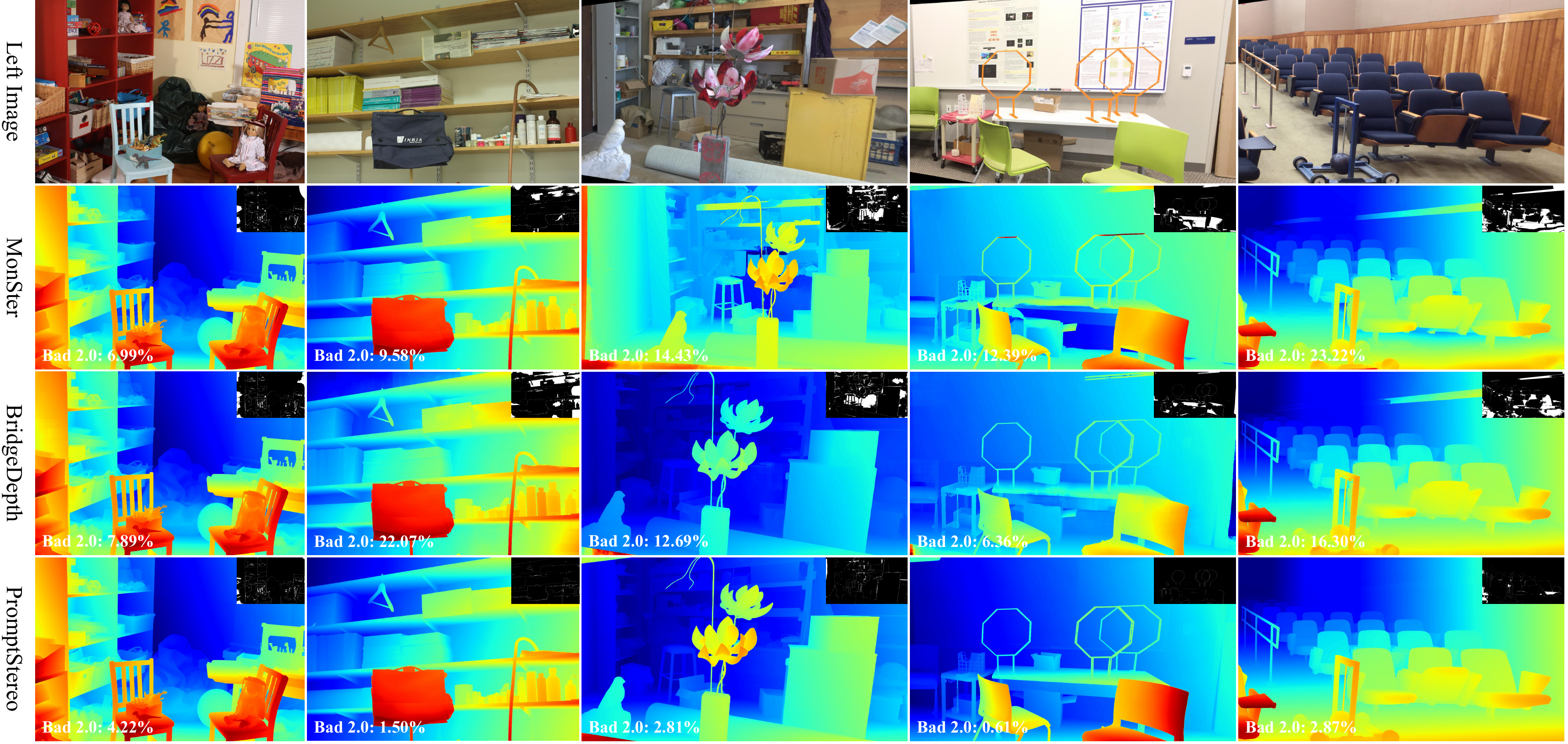}
    \caption{Visualization of Middlebury (unlimited training sets). The Bad 2.0 metric map is in the upper right corner.}
    \label{fig:middlebury}
    \vspace{-10px}
\end{figure*}

\subsection{Zero-Shot Generalization Benchmark}

We construct two zero-shot generalization benchmarks, namely basic and advanced. Each benchmark is further divided into two training settings: Scene Flow~\cite{mayer2016large} and unlimited, where the latter excludes samples from target datasets. All methods are evaluated using the same testing code.

As shown in Tab. ~\ref{tab:basic}, our PromptStereo achieves state-of-the-art performance on the basic benchmark under the Scene Flow training setting, ranking first on most metrics and within the top three across all. Compared to the baseline MonSter~\cite{cheng2025monster}, PromptStereo achieves a remarkable improvement, reducing the error on Midd-2021 by nearly 50\%. It demonstrates that PromptStereo maintains robust performance even under challenging real-world conditions, such as imperfect rectification in mobile-captured scenes.

Under the unlimited training setting, PromptStereo continues to achieve state-of-the-art performance. Notably, it achieves over 50\% improvement on Midd-2021 compared to MonSter~\cite{cheng2025monster} and BridgeDepth~\cite{guan2025bridgedepth}, and even surpasses FoundationStereo~\cite{wen2025foundationstereo} that we do not include in direct comparisons. As shown in Fig. ~\ref{fig:middlebury}, PromptStereo shows noticeable improvements in both details and texture-less regions.

As shown in Tab. ~\ref{tab:advanced}, our PromptStereo maintains state-of-the-art performance on the advanced benchmark under the Scene Flow training setting. Compared to the baseline MonSter~\cite{cheng2025monster}, PromptStereo consistently achieves impressive improvements. It can also be observed that on Booster, which contains reflective and transparent surfaces, training solely on Scene Flow is insufficient to demonstrate the model capacity and scalability provided by PRU.

\begin{table*}
    \centering
    \resizebox{\textwidth}{!}{
    \begin{tabular}{l|ccccccccccccc}
    \hline
    \multirow{3}{*}{Method} & \multicolumn{8}{c|}{DrivingStereo (H)} & \multicolumn{5}{c}{Booster (Q)} \\ \cline{2-14} 
     & \multicolumn{2}{c|}{Cloudy} & \multicolumn{2}{c|}{Foggy} & \multicolumn{2}{c|}{Rainy} & \multicolumn{2}{c|}{Sunny} & \multirow{2}{*}{EPE} & \multirow{2}{*}{Bad 2.0} & \multirow{2}{*}{Bad 4.0} & \multirow{2}{*}{Bad 6.0} & \multirow{2}{*}{Bad 8.0} \\
     & EPE & \multicolumn{1}{c|}{Bad 3.0} & EPE & \multicolumn{1}{c|}{Bad 3.0} & EPE & \multicolumn{1}{c|}{Bad 3.0} & EPE & \multicolumn{1}{c|}{Bad 3.0} &  &  &  &  &  \\ \hline
    \multicolumn{1}{l}{Training Set} & \multicolumn{13}{c}{Scene Flow} \\ \hline
    RAFT-Stereo~\cite{lipson2021raft} & 0.97 & \multicolumn{1}{c|}{3.58} & \snd 0.95 & \multicolumn{1}{c|}{\snd 2.63} & 1.80 & \multicolumn{1}{c|}{11.01} & 1.01 & \multicolumn{1}{c|}{3.97} & 3.95 & 16.36 & 12.57 & 10.73 & 9.44 \\
    IGEV-Stereo~\cite{xu2023iterative} & 1.11 & \multicolumn{1}{c|}{4.84} & 1.12 & \multicolumn{1}{c|}{4.58} & 2.34 & \multicolumn{1}{c|}{14.87} & 1.18 & \multicolumn{1}{c|}{5.05} & 4.01 & 16.15 & 12.66 & 11.06 & 10.02 \\
    DEFOM-Stereo~\cite{jiang2025defom} & 0.98 & \multicolumn{1}{c|}{3.59} & 1.01 & \multicolumn{1}{c|}{3.51} & 1.35 & \multicolumn{1}{c|}{11.04} & 0.98 & \multicolumn{1}{c|}{3.77} & 2.92 & \trd 12.77 & \trd 9.53 & 8.22 & 7.42 \\
    MonSter\footnotemark[1]~\cite{cheng2025monster} & 1.10 & \multicolumn{1}{c|}{5.41} & 1.38 & \multicolumn{1}{c|}{9.44} & 1.74 & \multicolumn{1}{c|}{17.59} & 1.12 & \multicolumn{1}{c|}{4.96} & 3.04 & 19.17 & 11.49 & 8.25 & 7.69 \\
    BridgeDepth~\cite{guan2025bridgedepth} & \trd 0.94 & \multicolumn{1}{c|}{\snd 2.98} & 1.07 & \multicolumn{1}{c|}{4.34} & \fst 1.19 & \multicolumn{1}{c|}{\snd 6.40} & 0.99 & \multicolumn{1}{c|}{\snd 3.40} & 3.13 & 14.74 & 9.71 & \trd 8.01 & 7.03 \\
    Great-IGEV~\cite{li2025global} & 1.11 & \multicolumn{1}{c|}{4.93} & 1.02 & \multicolumn{1}{c|}{3.36} & 1.96 & \multicolumn{1}{c|}{12.20} & 1.12 & \multicolumn{1}{c|}{5.43} & 4.40 & 18.63 & 14.31 & 12.35 & 11.08 \\
    MGStereo~\cite{yao2025diving} & \fst 0.92 & \multicolumn{1}{c|}{3.36} & \snd 0.94 & \multicolumn{1}{c|}{\trd 2.85} & \snd 1.29 & \multicolumn{1}{c|}{\trd 9.39} & \fst 0.94 & \multicolumn{1}{c|}{3.70} & \fst 2.14 & \fst 9.97 & \fst 6.84 & \fst 5.75 & \fst 5.23 \\
    SMoE-Stereo~\cite{wang2025learning} & \snd 0.93 & \multicolumn{1}{c|}{\trd 3.12} & 1.11 & \multicolumn{1}{c|}{4.77} & \trd 1.31 & \multicolumn{1}{c|}{\fst 6.08} & 1.00 & \multicolumn{1}{c|}{\trd 3.51} & \trd 2.80 & 20.81 & 11.80 & 8.65 & \trd 6.85 \\
    PromptStereo (Ours) & \fst 0.92 & \multicolumn{1}{c|}{\fst 2.86} & \fst 0.89 & \multicolumn{1}{c|}{\fst 2.43} & 1.59 & \multicolumn{1}{c|}{14.30} & \fst 0.94 & \multicolumn{1}{c|}{\fst 3.16} & \snd 2.78 & \snd 12.13 & \snd 8.93 & \snd 7.54 & \snd 6.79 \\ \hline
    \multicolumn{1}{l}{Training Set} & \multicolumn{13}{c}{Unlimited} \\ \hline
    FoundationStereo\footnotemark[2]~\cite{wen2025foundationstereo} & 0.91 & \multicolumn{1}{c|}{2.94} & 0.92 & \multicolumn{1}{c|}{2.52} & 1.44 & \multicolumn{1}{c|}{11.26} & 0.94 & \multicolumn{1}{c|}{3.50} & 1.23 & 5.15 & 3.87 & 3.39 & 3.03 \\ \hdashline
    MonSter~\cite{cheng2025monster} & \trd 0.99 & \multicolumn{1}{c|}{\trd 3.28} & \trd 1.15 & \multicolumn{1}{c|}{\trd 5.30} & \snd 1.15 & \multicolumn{1}{c|}{\snd 5.36} & \snd 1.03 & \multicolumn{1}{c|}{\snd 3.66} & \trd 1.76 & 11.55 & 7.16 & 5.49 & \trd 4.49 \\
    BridgeDepth~\cite{guan2025bridgedepth} & \snd 0.95 & \multicolumn{1}{c|}{\snd 3.10} & \snd 1.10 & \multicolumn{1}{c|}{\snd 4.77} & \trd 1.32 & \multicolumn{1}{c|}{\trd 6.71} & \snd 0.98 & \multicolumn{1}{c|}{\snd 3.59} & 1.96 & \trd 11.25 & \trd 6.95 & \trd 5.52 & 4.57 \\
    MGStereo\footnotemark[3]~\cite{yao2025diving} & - & \multicolumn{1}{c|}{-} & - & \multicolumn{1}{c|}{-} & - & \multicolumn{1}{c|}{-} & - & \multicolumn{1}{c|}{-} & \snd 1.24 & \snd 7.91 & \snd 5.97 & \snd 4.08 & \snd 3.44 \\
    PromptStereo (Ours) & \fst 0.92 & \multicolumn{1}{c|}{\fst 3.09} & \fst 0.94 & \multicolumn{1}{c|}{\fst 2.71} & \trd 1.39 & \multicolumn{1}{c|}{\trd 10.25} & \fst 0.90 & \multicolumn{1}{c|}{\fst 3.29} & \fst 0.67 & \fst 3.67 & \fst 2.35 & \fst 1.95 & \fst 1.70 \\ \hline
    \end{tabular}
    }
    \vspace{-5px}
    \caption{Zero-shot generalization advanced benchmark. MonSter$^*$ is re-trained with the official code. FoundationStereo$^{\dagger}$ is reported for reference only, as it requires significantly large batch sizes and model parameters to train, making a fair comparison infeasible. MGStereo$^{\ddagger}$ is reported from the official GitHub repository, which is fine-tuned on TranScene~\cite{liu2025multi} without available checkpoints.}
    \label{tab:advanced}
    \vspace{-5px}
\end{table*}

\begin{figure*}[t]
    \centering
    \includegraphics[width=1.0\textwidth]{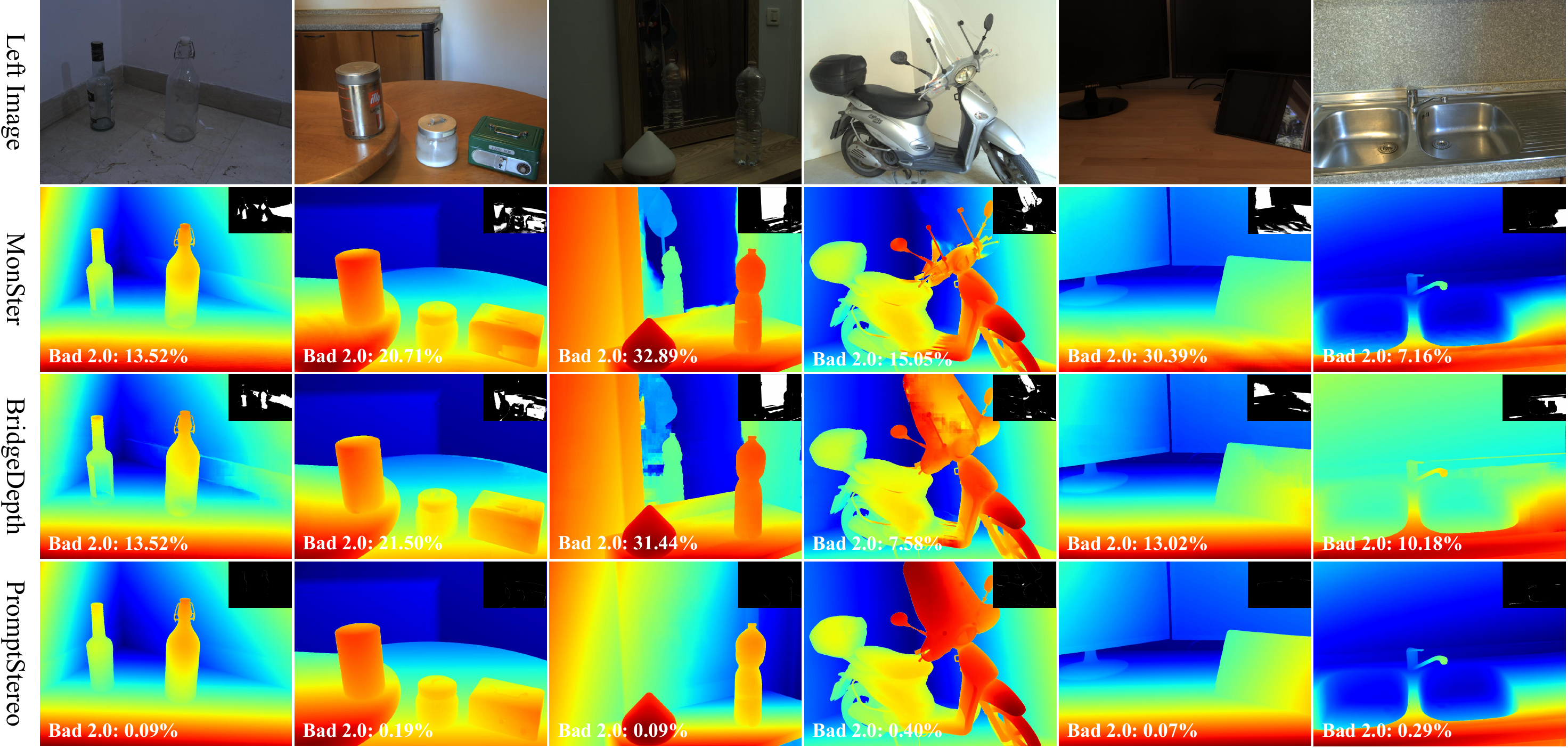}
    \caption{Visualization of Booster (unlimited training sets). The Bad 2.0 metric map is in the upper right corner.}
    \label{fig:booster}
    \vspace{-10px}
\end{figure*}

However, under the unlimited training setting, PromptStereo achieves remarkable improvements on Booster, surpassing the second-best method, MGStereo\footnotemark[3], by over 50\%. Notably, MGStereo\footnotemark[3] is fine-tuned on TransScene~\cite{liu2025multi}. Furthermore, PromptStereo also outperforms FoundationStereo~\cite{wen2025foundationstereo}, highlighting the strong representation capacity and scalability of PRU. As shown in Fig. ~\ref{fig:booster}, our PromptStereo performs well on the most challenging scenarios. It accurately predicts disparities even in regions with screens, reflective materials, and transparent surfaces, demonstrating that PRU effectively leverages monocular depth priors.

\subsection{Ablation Study}

We begin to study the contribution of each proposed component. For efficiency, our ablation studies may use reduced data augmentations, smaller crop size, or fewer training iterations. Nevertheless, all studies within the same group share identical training configurations, ensuring a fair comparison. Our ablation studies cover four aspects: module effectiveness, module universality, PRU design, and iteration number. Through these studies, we aim to provide a deeper understanding of how PRU leverages prompts and holds superior representation capacity and scalability.

\begin{table*}
    \centering
    \begin{tabular}{l|ccc|cc|cc|c}
    \hline
    \multirow{2}{*}{Model} & \multirow{2}{*}{MP} & \multirow{2}{*}{SP} & \multirow{2}{*}{AIF} & \multicolumn{2}{c|}{Midd-T (H)} & \multicolumn{2}{c|}{ETH3D} & \multirow{2}{*}{Time (s)} \\
     &  &  &  & EPE & Bad 2.0 & EPE & Bad 1.0 &  \\ \hline
    Baseline (MonSter) &  &  &  & 0.94 & 7.27 & 0.26 & 2.86 & 0.64 \\ \hline
    PRU + MP & \checkmark &  &  & 0.66 & 4.18 & 0.22 & 1.38 & 0.35 \\
    PRU + MP + SP & \checkmark & \checkmark &  & 0.62 & 3.90 & 0.21 & 1.35 & 0.36 \\
    Full model (PromptStereo) & \checkmark & \checkmark & \checkmark & \textbf{0.60} & \textbf{3.76} & \textbf{0.20} & \textbf{1.30} & 0.36 \\ \hline
    \end{tabular}
    \vspace{-5px}
    \caption{Ablation study of module effectiveness. Time is recorded when inferencing Scene Flow.}
    \label{tab:effectiveness}
    \vspace{-5px}
\end{table*}

\begin{table} \footnotesize
    \centering
    \begin{tabular}{l|c|c|c|c}
    \hline
    \multirow{2}{*}{Model} & KITTI 2015 & Midd-T (H) & ETH3D & \multirow{2}{*}{Time (s)} \\
     & Bad 3.0 & Bad 2.0 & Bad 1.0 &  \\ \hline
    RAFT-Stereo & 5.68 & 8.41 & 2.29 & 0.36 \\
    Prompt-RAFT & \textbf{4.78} & \textbf{6.39} & \textbf{1.49} & 0.38 \\ \hline
    IGEV-Stereo & 6.03 & 7.04 & 3.61 & 0.37 \\
    Prompt-IGEV & \textbf{4.84} & \textbf{6.50} & \textbf{2.21} & 0.38 \\ \hline
    MonSter & 5.52 & 7.27 & 2.86 & 0.64 \\
    PromptStereo & \textbf{4.59} & \textbf{3.76} & \textbf{1.30} & 0.36 \\ \hline
    \end{tabular}
    \vspace{-5px}
    \caption{Ablation study of module universality.}
    \label{tab:universality}
    \vspace{-5px}
\end{table}

\begin{table} \footnotesize
    \centering
    \begin{tabular}{l|c|c|c}
    \hline
    \multirow{2}{*}{Model} & KITTI 2015 & Midd-T (H) & ETH3D \\
     & Bad 3.0 & Bad 2.0 & Bad 1.0 \\ \hline
    w/o pre-trained weights & 5.03 & 4.39 & 1.39 \\
    w/ hidden state conv & 4.86 & 4.37 & 1.34 \\
    w/ zero-init conv & 4.83 & 4.04 & 1.42 \\
    PromptStereo & \textbf{4.59} & \textbf{3.76} & \textbf{1.30} \\ \hline
    \end{tabular}
    \vspace{-5px}
    \caption{Ablation study of PRU design.}
    \label{tab:pru}
    \vspace{-5px}
\end{table}

\textbf{Module Effectiveness.} As shown in Tab. ~\ref{tab:effectiveness}, we take MonSter~\cite{cheng2025monster} as our baseline. Replacing its dual-branch GRU with our PRU + MP already brings superior improvements in both inference speed and accuracy. With SP and AIF further incorporated, performance consistently increases, while inference time grows only marginally. Our PRU unifies monocular and stereo cues within a single branch, achieving gains in both efficiency and accuracy.

\textbf{Module Universality.} As shown in Tab. ~\ref{tab:universality}, we replace GRU in RAFT-Stereo~\cite{lipson2021raft} and IGEV-Stereo~\cite{xu2023iterative} with PRU to evaluate its universality. The results show that PRU improves accuracy while maintaining nearly unchanged inference time. It demonstrates that PRU serves as a generic and pluggable recurrent unit, improving the performance of iterative stereo matching while preserving efficiency.

\textbf{PRU Design.} As shown in Tab. ~\ref{tab:pru}, when we remove the pre-trained weights, it leads to a clear performance drop, confirming the benefit of inheriting priors. Interestingly, even with random initialization, our model still achieves a competitive performance, suggesting that PRU's architecture is also effective. When we merge the hidden state and prompt into a single convolutional block, performance decreases, showing that such coupling disturbs the separation of information. Additionally, replacing the last convolutional layer of the prompt block with a zero-initialized convolution~\cite{lin2025prompting} also degrades performance, indicating that this design hinders training convergence when the prompt information is sufficiently rich.  

\begin{table} \footnotesize
    \centering
    \begin{tabular}{l|ccccc}
    \hline
    \multirow{2}{*}{Model} & \multicolumn{5}{c}{Iteration Number} \\
     & 4 & 8 & 12 & 16 & 32 \\ \hline
    \multicolumn{1}{l}{Evaluation Set} & \multicolumn{5}{c}{Midd-2021} \\ \hline
    MonSter & 10.75 & 10.64 & 10.16 & 9.61 & 8.46 \\
    PromptStereo & \textbf{4.35} & \textbf{3.28} & \textbf{2.93} & \textbf{2.79} & \textbf{2.78} \\ \hline
    \multicolumn{1}{l}{Evaluation Set} & \multicolumn{5}{c}{Booster (Q)} \\ \hline
    MonSter & 14.09 & 13.62 & 12.82 & 12.62 & 11.55 \\
    PromptStereo & \textbf{5.25} & \textbf{4.39} & \textbf{4.24} & \textbf{4.09} & \textbf{3.67} \\ \hline
    \end{tabular}
    \vspace{-5px}
    \caption{Ablation study of iteration number.}
    \label{tab:iteration}
    \vspace{-5px}
\end{table}

\textbf{Iteration Number.} As shown in Tab. ~\ref{tab:iteration}, we evaluate the impact of iteration numbers on Midd-2021 and Booster. It demonstrates that our PromptStereo converges significantly faster and achieves higher accuracy under the same number of iterations. On Midd-2021, PromptStereo achieves near-optimal performance within 16 iterations, whereas on Booster, performance continues to improve up to 32 iterations, which is expected given the difficulty of reflective and transparent surfaces. These results further validate the improved convergence speed brought by PRU.
\section{Conclusion}
\label{sec:conclusion}

We propose PromptStereo, a novel iterative stereo matching method for zero-shot generalization. It replaces GRU with Prompt Recurrent Unit (PRU), which inherits monocular depth priors and offers better representation capacity and scalability. Structure Prompt (SP) and Motion Prompt (MP) prompts monocular structure and stereo motion cues into PRU, providing clear guidance and maintaining independent state information. We further utilize Affine-Invariant Fusion (AIF) for global geometric consistent initialization, and introduce a simple yet effective update strategy for efficient iterative refinement. Experiments demonstrate that our PromptStereo achieves state-of-the-art zero-shot performance across various datasets, including challenging real-world scenarios. Our work highlights the potential of prompt-guided iterative refinement in stereo matching. 

\textbf{Limitations.} Despite its strong zero-shot generalization, PromptStereo's performance in extreme adverse weather scenarios is still constrained. There may be various reasons that deserve further investigation in the future.

\textbf{Acknowledgement.} This research is supported by the National Natural Science Foundation of China (62472184, 62122029) and the Fundamental Research Funds for the Central Universities.

{
    \small
    \bibliographystyle{ieeenat_fullname}
    \bibliography{main}
}

\clearpage
\setcounter{page}{1}
\maketitlesupplementary

\begin{figure}[t]
    \centering
    \includegraphics[width=1.0\linewidth]{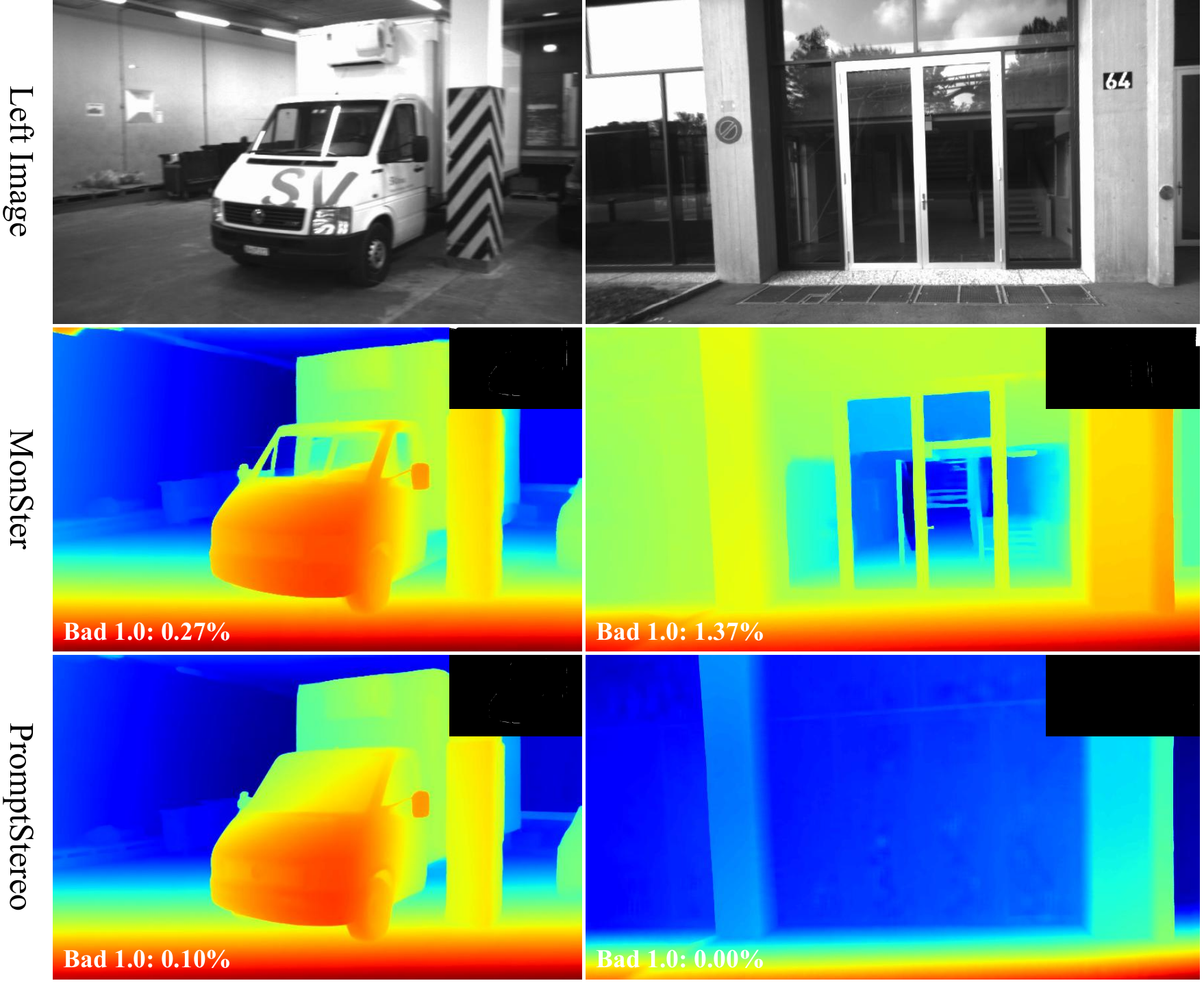}
    \caption{Visualization of ETH3D (unlimited training sets). The Bad 1.0 metric map is in the upper right corner.}
    \label{fig:eth3d}
    \vspace{-5px}
\end{figure}

\begin{table}
    \centering
    \resizebox{\linewidth}{!}{
        \begin{tabular}{l|c|c|c|c|c}
        \hline
        \multirow{2}{*}{Model} & KITTI 2012 & KITTI 2015 & Midd-T (H) & Midd-2021 & ETH3D \\
         & Bad 3.0 & Bad 3.0 & Bad 2.0 & Bad 2.0 & Bad 1.0 \\ \hline
        MonSter & \textbf{3.27} & 3.72 & 3.75 & 8.46 & 1.02 \\
        Ours w/o FSD & 3.34 & 3.50 & 2.41 & 3.97 & \textbf{0.71} \\
        Ours w/ FSD & 3.33 & \textbf{3.40} & \textbf{2.21} & \textbf{2.78} & 0.79 \\ \hline
        \end{tabular}
    }
    \vspace{-5px}
    \caption{Ablation study of whether to use FSD~\cite{wen2025foundationstereo}.}
    \label{tab:dataset}
    \vspace{-5px}
\end{table}

\section{Additional Experiment}
\label{sec:supp}

In this section, we provide additional comparisons with several methods, including implementation details and method-specific training strategies. We compare our PromptStereo with three methods: MonSter~\cite{cheng2025monster}, FoundationStereo~\cite{wen2025foundationstereo}, and Stereo Anywhere~\cite{bartolomei2025stereo}.

\subsection{MonSter}

In the main text, we evaluate MonSter~\cite{cheng2025monster} with a re-trained Scene Flow checkpoint. Although the official repository provides a Scene Flow checkpoint, our evaluation reveals inconsistent results. The public checkpoint shows unrealistically strong performance. After consulting the authors, we have been informed that the public Scene Flow checkpoint is not trained solely on Scene Flow. They uploaded the wrong checkpoint, and they plan to correct it in the future. Therefore, we re-train MonSter using the official code.

As shown in Fig. ~\ref{fig:eth3d}, we further visualize results on ETH3D. Although it does not provide ground truth for glass surfaces, the visualization clearly shows that our PromptStereo correctly infers these regions even on grayscale images. At the same time, MonSter fails to handle such challenging transparent areas.

Besides, we investigate the influence of training sets. Since MonSter's mixed training sets do not include FSD~\cite{wen2025foundationstereo}, we remove FSD and add TartanAir, following MonSter's configuration. As shown in Tab. ~\ref{tab:dataset}, excluding FSD leads to only a marginal decrease in accuracy (except ETH3D), and our PromptStereo still surpasses MonSter. This further demonstrates the effectiveness and universality of PromptStereo.

\subsection{FoundationStereo}

In the main text, we report FoundationStereo's results, but do not include it in our comparison. This is due to two main practical constraints. First, FoundationStereo~\cite{wen2025foundationstereo} requires a tremendous training configuration (a batch size of 128 on 32 A100 GPUs), whereas other methods only use a batch size of 8 on 4 GPUs with 24 GB of memory. Second, only the inference code is publicly available; the training code has not been released, making its results difficult to reproduce. For these reasons, we exclude FoundationStereo from direct comparisons in the main text.

\begin{figure*}[t]
    \centering
    \includegraphics[width=1.0\textwidth]{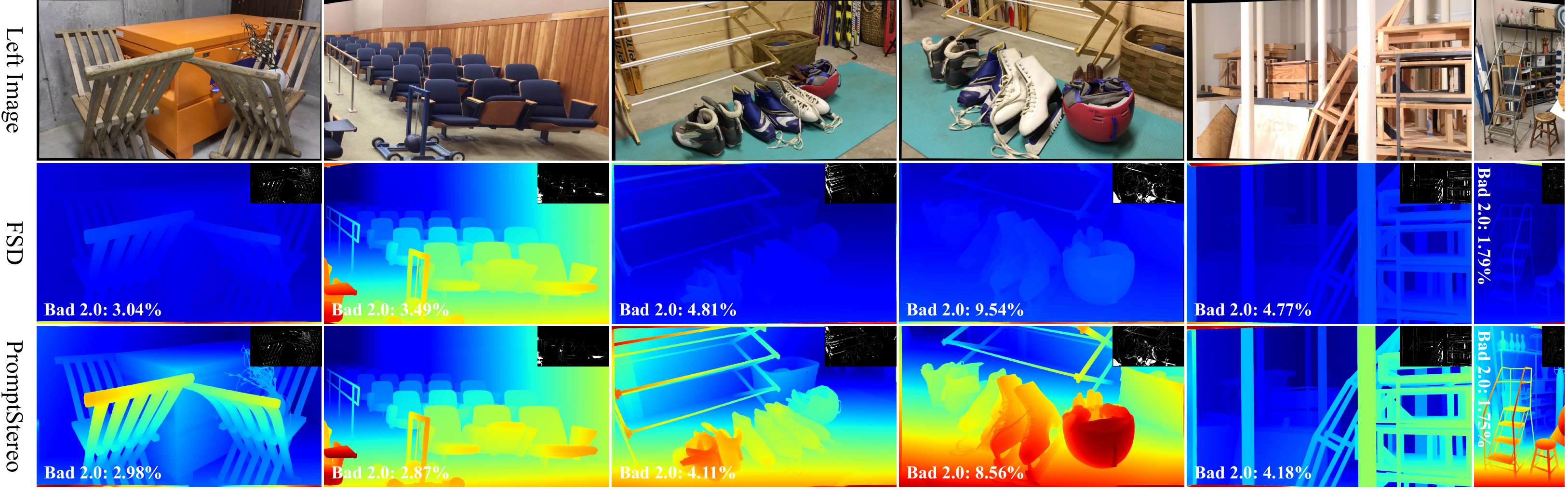}
    \caption{Visualization of Middlebury 2021 (unlimited training sets). The Bad 2.0 metric map is in the upper right corner.}
    \label{fig:supp_middlebury}
    \vspace{-5px}
\end{figure*}

\begin{figure*}[t]
    \centering
    \includegraphics[width=1.0\textwidth]{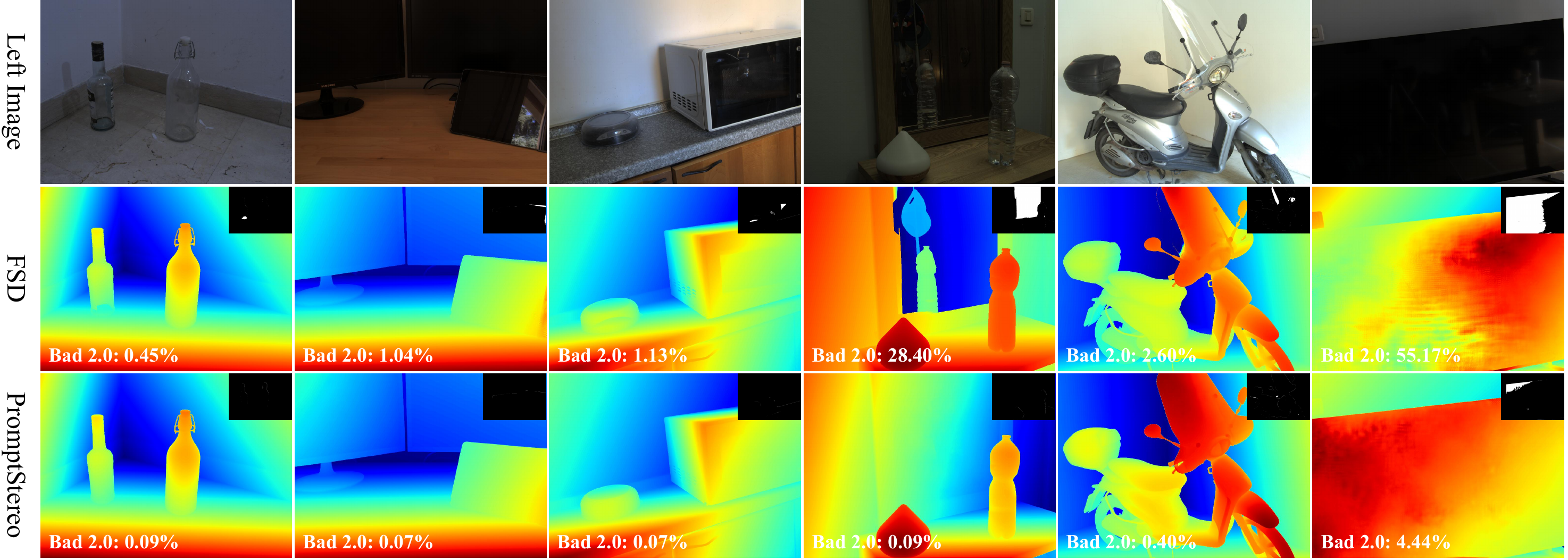}
    \caption{Visualization of Booster (unlimited training sets). The Bad 2.0 metric map is in the upper right corner.}
    \label{fig:supp_booster}
    \vspace{-10px}
\end{figure*}

However, our PromptStereo still surpasses FoundationStereo on Middlebury 2021 and Booster. As shown in Fig. ~\ref{fig:supp_middlebury}, PromptStereo achieves slightly better accuracy on Middlebury 2021. Interestingly, FoundationStereo produces extreme and unstable disparity estimates along the image borders, which in turn shift the overall color visualization, whereas PromptStereo maintains well-behaved predictions and fine color visualization. Moreover, as shown in Fig. ~\ref{fig:supp_booster}, FoundationStereo remains poor performance on reflective and transparent surfaces, whereas PromptStereo can provide reasonable predictions even on large mirror surfaces. This robustness to transparent regions is a key advantage of PromptStereo over FoundationStereo.

\subsection{Stereo Anywhere}

Stereo Anywhere is also a special case. Its main innovations lie in its normal volume construction, volume truncation, and volume augmentation. Beyond these design aspects, our inspection of the released code shows two additional factors that may affect the fairness of comparison. First, its data augmentation differs from RAFT-Stereo's standard augmentation, which includes more than a dozen transformations (ChannelDropout, ChannelShuffle, MotionBlur, ImageCompression, GaussNoise, etc). Second, Stereo Anywhere is not trained from scratch; instead, it initializes training from a RAFT-Stereo checkpoint. To make the comparison fair, we re-train PromptStereo on Scene Flow from a PromptStereo Scene Flow checkpoint and apply the same data augmentation as Stereo Anywhere.

\begin{table}
    \centering
    \resizebox{\linewidth}{!}{
        \begin{tabular}{l|c|c|c|c|c}
        \hline
        \multirow{2}{*}{Model} & KITTI 2012 & KITTI 2015 & Midd-T (H) & Midd-2021 & ETH3D \\
         & Bad 3.0 & Bad 3.0 & Bad 2.0 & Bad 2.0 & Bad 1.0 \\ \hline
        Stereo Anywhere & 3.90 & \textbf{3.93} & 4.49 & 5.18 & 1.43 \\
        Ours & 3.77 & 4.59 & \textbf{3.76} & \textbf{4.84} & 1.30 \\
        Ours w/ STA & \textbf{3.33} & 4.12 & 4.03 & 5.06 & \textbf{1.26} \\ \hline
        \end{tabular}
    }
    \vspace{-5px}
    \caption{Ablation study of special training strategy.}
    \label{tab:augmentation}
    \vspace{-5px}
\end{table}

As shown in Tab. ~\ref{tab:augmentation}, the re-trained PromptStereo model achieves improvements on KITTI and ETH3D, with notable gains on KITTI. Performance on Middlebury decreases slightly, yet both versions of PromptStereo still outperform Stereo Anywhere. We also measure inference time on Scene Flow: Stereo Anywhere requires 0.65 seconds per sample, while PromptStereo needs only 0.36 seconds, representing a substantial efficiency advantage.

\end{document}